\documentclass[conference]{IEEEtran}

\IEEEoverridecommandlockouts
\usepackage{lipsum}
\usepackage{pdfx}
\hypersetup{
    pdfstartview={FitH},
    pdfinfo={
        Title={Your Title},
        Author={Your Name},
        Subject={},
        Keywords={}
    }
}

\pdfmapline{=phvr8r Helvetica-Regular <8r.enc}
\pdfmapline{=phvb8r Helvetica-Bold <8r.enc}
\pdfmapline{=phvro8r Helvetica-Oblique <8r.enc}
\pdfmapline{=phvbo8r Helvetica-BoldOblique <8r.enc}

\usepackage{cite}
\usepackage{amsmath,amssymb,amsfonts}
\usepackage{algorithmic}
\usepackage{graphicx}
\usepackage{textcomp}
\usepackage{helvet}
\usepackage{multirow}
\usepackage{pdfpages}
\usepackage{booktabs}
 \usepackage{graphicx}

\def\BibTeX{{\rm B\kern-.05em{\sc i\kern-.025em b}\kern-.08em
    T\kern-.1667em\lower.7ex\hbox{E}\kern-.125emX}}
\begin{document}

\title{FH-SSTNet: Forehead Creases based User Verification using Spatio-Spatial Temporal Network \\
}

\author{Geetanjali Sharma\textsuperscript{1} \quad \quad   Gaurav Jaswal \textsuperscript{2} \quad \quad   Aditya Nigam \textsuperscript{1} \quad \quad Raghavendra Ramachandra\textsuperscript{3}\\
\textsuperscript{1}Indian Institute of Technology Mandi, India \\
\textsuperscript{2}Technology Innovation Hub- Indian Institute of Technology, Mandi\\
\textsuperscript{3}Norwegian University of Science and Technology (NTNU), Norway.
}


\IEEEoverridecommandlockouts \IEEEpubid{\makebox[\columnwidth]{TBA \hfill} \hspace{\columnsep}\makebox[\columnwidth]{ }}

\maketitle
\IEEEpubidadjcol

\begin{abstract}
Biometric authentication, which utilizes contactless features, such as forehead patterns, has become increasingly important for identity verification and access management. The proposed method is based on learning a 3D spatio-spatial temporal convolution to create detailed pictures of forehead patterns. We introduce a new CNN model called the Forehead Spatio-Spatial Temporal Network (FH-SSTNet), which utilizes a 3D CNN architecture with triplet loss to capture distinguishing features. We enhance the model's discrimination capability using Arcloss in the network's head. Experimentation on the Forehead Creases version 1 (FH-V1) dataset, containing 247 unique subjects, demonstrates the superior performance of FH-SSTNet compared to existing methods and pre-trained CNNs like ResNet50, especially for forehead-based user verification. The results demonstrate the superior performance of FH-SSTNet for forehead-based user verification, confirming its effectiveness in identity authentication.

\end{abstract}

\begin{IEEEkeywords}
Biometrics, Forehead creases, Spatial-Spatio Temporal, metric learning, Verification.
\end{IEEEkeywords}

\section{Introduction}
Biometric traits are of paramount importance in modern verification systems and provide secure authentication across a wide range of access control applications In contrast to traditional methods, such as passwords, biometrics offer reliable verification due to their distinctive characteristics, thereby enhancing the security and privacy of individuals. The benefits of biometric authentication extend beyond security and include convenience and ease of use.


\begin{figure}[htp]
\centering
\includegraphics[width=1\columnwidth]{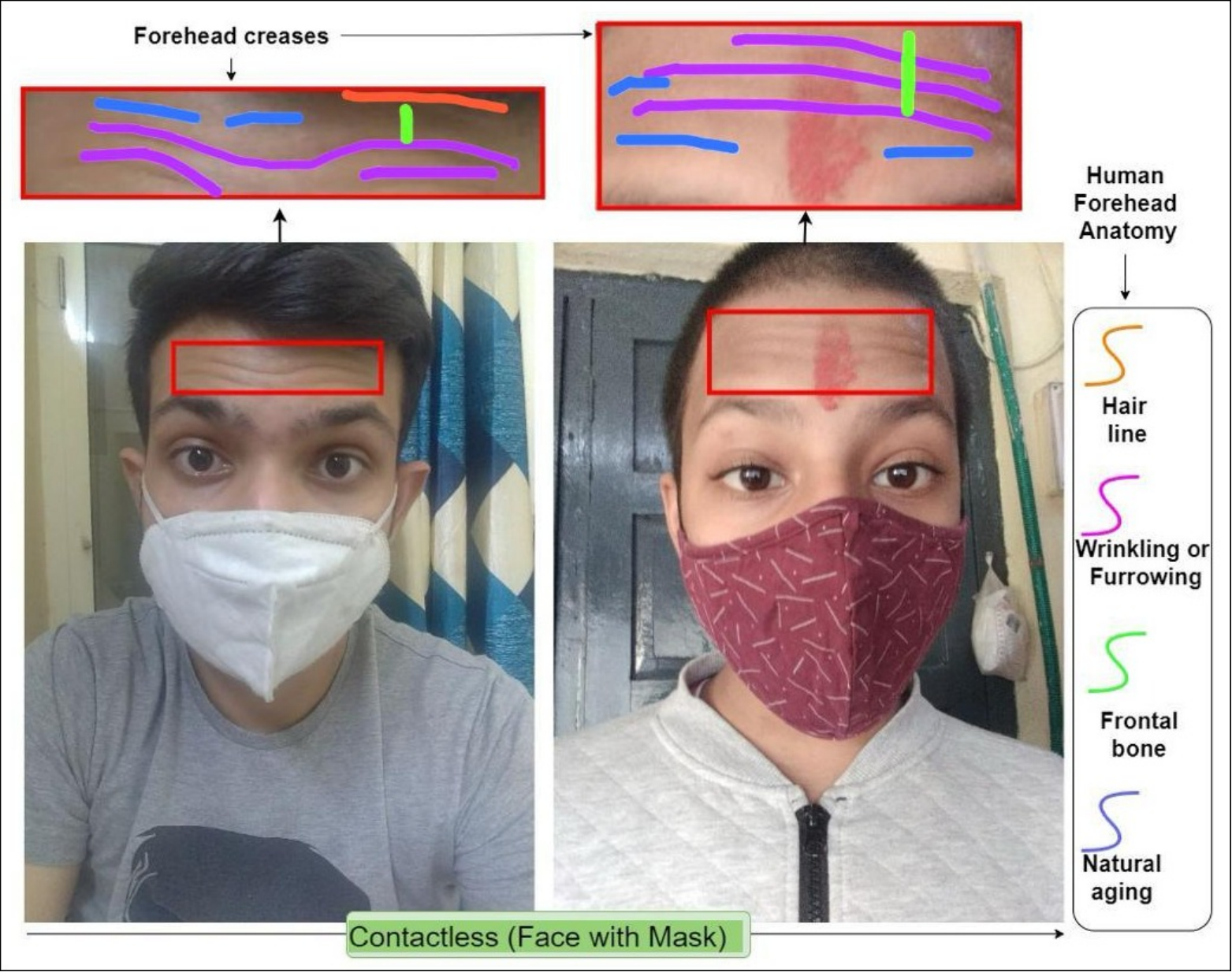}
\caption{Forehead creases are characterized by the presence of vertical, horizontal, and diagonal lines under facial expression, which serve as boundary markers along with the textured patterns observed on the skin's surface. In human forehead anatomy, forehead creases result from their connection to the muscles below the eyelids, Frontal muscles create facial expressions, like raising eyebrows and wrinkling the forehead, Frontal-is muscles meet in the middle, causing forehead wrinkles and Age, Sun exposure, and facial expression cause noticeable lines.}
\label{fig:figure1}
\end{figure}

Biometric authentication encompasses a diverse range of modalities, each of which has unique characteristics and applications. From fingerprints \cite{dong2022voiceprint} and iris \cite{daugman1993high}\cite{daugman2009iris} patterns to facial \cite{fu2019dual} recognition and voice prints \cite{dong2022voiceprint}, palmprints \cite{thapar2019pvsnet} biometric modalities offer versatile options for authentication, catering to varying user preferences and system requirements. This diversity enables organizations to implement authentication solutions tailored to their specific needs, whether it enhances security at access points or streamlines user (or person) authentication for digital transactions. Furthermore, certain biometric traits, such as finger-veins \cite{8672597}, exhibit remarkable resilience to spoofing attempts while maintaining stability over time. This feature, coupled with the easy accessibility for biometric authentication, further enhances the reliability and effectiveness of biometric systems.  The utility of the particular biometric characteristics depends on the use case  and the application that determines the trade off between security, convenience, and accuracy.

Although traditional biometric characteristics are extensively utilized in various access control applications, their performance can be hindered by unconstrained capture conditions. For instance, the use of a surgical face mask and eye accessories, such as eyeglasses, can hinder the performance of facial, periocular, or iris biometrics. In such cases, the use of soft biometric traits such as forehead creases can offer a unique and potentially valuable biometric trait for user verification. Forehead creases are created by the movement of muscles, which can be attributed to facial expression and their unique features can include a set of horizontal, vertical, and diagonal edges, as well as textured patterns in the skin region. Figure \ref{fig:figure1} illustrates an example of the utility of forehead creases that can be captured using the frontal camera from a smartphone when the data subject is wearing surgical masks and eye accessories.


The use of forehead biometric characteristics is relatively new in biometrics research. As a result, only a limited amount of related research is available in the literature. The first work on forehead recognition was presented in \cite{manit2020human}. This initial study acquired forehead images using a complex imaging system based on near-infrared laser scanning. However, the setup used in this study was not suitable for smartphone-based applications because of hygiene concerns, sensor surface noise, and user inconvenience. The approach detailed in this reference served as a baseline without many optimizations for smartphone-based user recognition. The utility of forehead creases as a biometric characteristic was first introduced by Bharadwaj et al. \cite{bharadwaj2022mobile}. Their work not only proposed the application and evaluation under COVID-19 masked face scenarios, but also introduced a new approach to employing forehead creases as a biometric method for identifying individuals in the face of the challenges posed by the pandemic. 

This work introduces the innovative concept of utilizing forehead creases as a unique biometric trait for user recognition, particularly in scenarios where traditional identifiers, such as facial and periocular features, are obscured by face masks. Forehead crease patterns, which are distinct and enduring, offer promise for reliable biometric recognition. To facilitate this, the researchers compiled a dataset comprising 4,964 image samples from 247 subjects across two sessions, capturing various facial expressions and eyebrow movements to reveal the forehead wrinkles. The dataset aims to ensure temporal stability and is collected remotely using a mobile application in a natural setting. The proposed forehead recognition system encompasses several stages, including forehead Region of Interest (RoI) segmentation, discriminative feature extraction, and matching. Yolo-v3 \cite{adarsh2020yolo} was utilized for segmentation, whereas ResNet18 \cite{targ2016resnet}, augmented with a dual-attention mechanism, served as the backbone network. Large Margin Cosine Loss (LMCL) is adopted as the loss function to enhance feature discrimination. Deep metric learning techniques, including LMCL and dual attention mechanisms, were employed to extract discriminative features from forehead images. Evaluation of the proposed framework involves assessing performance metrics such as True Match Rate (TMR) @ False Match Rate (FMR) and  Equal Error Rate (EER) in verification scenarios. 


\subsection{Open challenges and our work}
Our objective differs from that of typical CNN-based biometric studies such as ~\cite{schroff2015facenet, deng2019arcface, thapar2019pvsnet}. Instead of creating a new recognition system, we explore whether 3D-CNNs can autonomously discern the importance of nonrigid motion in biometric matching. We aim to exploit CNNs' feature learning capability of CNNs to evaluate their alignment with human perception of the significance of non-rigid motion across various poses and lighting conditions. Unlike existing methods that use plain convolution filters, we seek more comprehensive feature extraction. Additionally, we aim to streamline the triplet loss-based model by simplifying the triplet selection during training.

In our work, we propose a deep metric learning-based DNN framework tailored for effective matching across contactless forehead crease datasets, addressing image-specific challenges. Our DNN architecture incorporates a 3DCNN network, leveraging its ability to extract highly discriminative spatio-spatial temporal features invariant to rotation, scale, illumination, and occlusion. By stacking multiple patches with overlapping information, we input a sequence of patches to the 3DCNN\cite{klaiber2021systematic}, maximizing the utilization of rectangular filters and enhancing the intra-patch spatio-spatial temporal features. Dual-loss training further refines the inter-patch temporal dependencies, ensuring robust performance. This end-to-end trainable CNN is augmented with triplet loss and Arc-Face, facilitating the generalized learning of unique embeddings in the transformation space across diverse contactless iris and forehead datasets.
 In summary, our paper contributes threefold:
\begin{itemize}
    \item To the best of the author's knowledge, this is the first study that learns the spatio-spatial temporal features for forehead-based person verification.
    \item We proposed a  3D-CNN  architecture called FH-SSTNet using rectangular and well-designed 3D convolution, which can generate a high discriminative intra-patch spatio-spatial temporal features.
        \item Extensive experiments have been conducted on the Forehead Creases version 1 (FH-V1) dataset. The performance of the proposed method was compared to that of existing methods, and the results obtained indicated that the proposed FH-SSTNet achieved the best performance among all the methods.

\end{itemize}

The rest of the paper is orgnaised as follows: Section \ref{sec:Pro} introduce the proposed method, Section \ref{sec:Exp} experiment and results and Section \ref{sec:Cocn} draws the conclusion. 
\section{Proposed Methodology}
\label{sec:Pro}
In this section, we propose a framework for user verification that utilizes forehead features, as depicted in Figure \ref{fig:figure3}. The novelty of the proposed approach is the use of spatio-spatial temporal features extracted from the forehead of the user. Furthermore, we introduce a new 3D CNN architecture inspired by inception-v1 \cite{szegedy2015going} which is trained end-to-end using triplet loss to effectively capture the  spatio-spatial temporal features. Finally, we proposed a head (or predictor) that processes the features using two fully connected layers that are trained using Arcloss to obtain the final feature embedding of dimension $1 \times 512$. Finally, cosine distance is used to compute the verification score. We named the proposed method the ForeHead-Spatio-Spatial Temporal Network (FH-SSTNet). In the following sections, we discuss functional blocks of the proposed FH-SSTNet such as  image montages cube and FH-SSTNet(backbone and head).  
Owing to the sequential stacking of biometric trait images, composed of separate small patches, the individual elements within these patches become interconnected, indicating generalized image sequential stacking in 3D. They contribute to learning spatio-spatial temporal features by capturing and processing information related to both the spatial and temporal dimensions. In the context of spatio-spatial temporal features in image data or sequences, "spatio-spatial" refers to the spatial (2D) aspects, while "temporal" pertains to the temporal (time-based) aspects. 

\subsection{Image Montage}
The process of dividing images into small patches serves as the foundation for the model analysis of local details and the capture of fine-grained spatial features within each patch. This approach establishes a basis for understanding intricate spatial characteristics. Moreover, the interconnections between these patches play a crucial role in enabling the model to identify relationships and patterns among neighboring spatial regions.

\begin{figure}[htp]
\centering
\includegraphics[width=1\columnwidth]{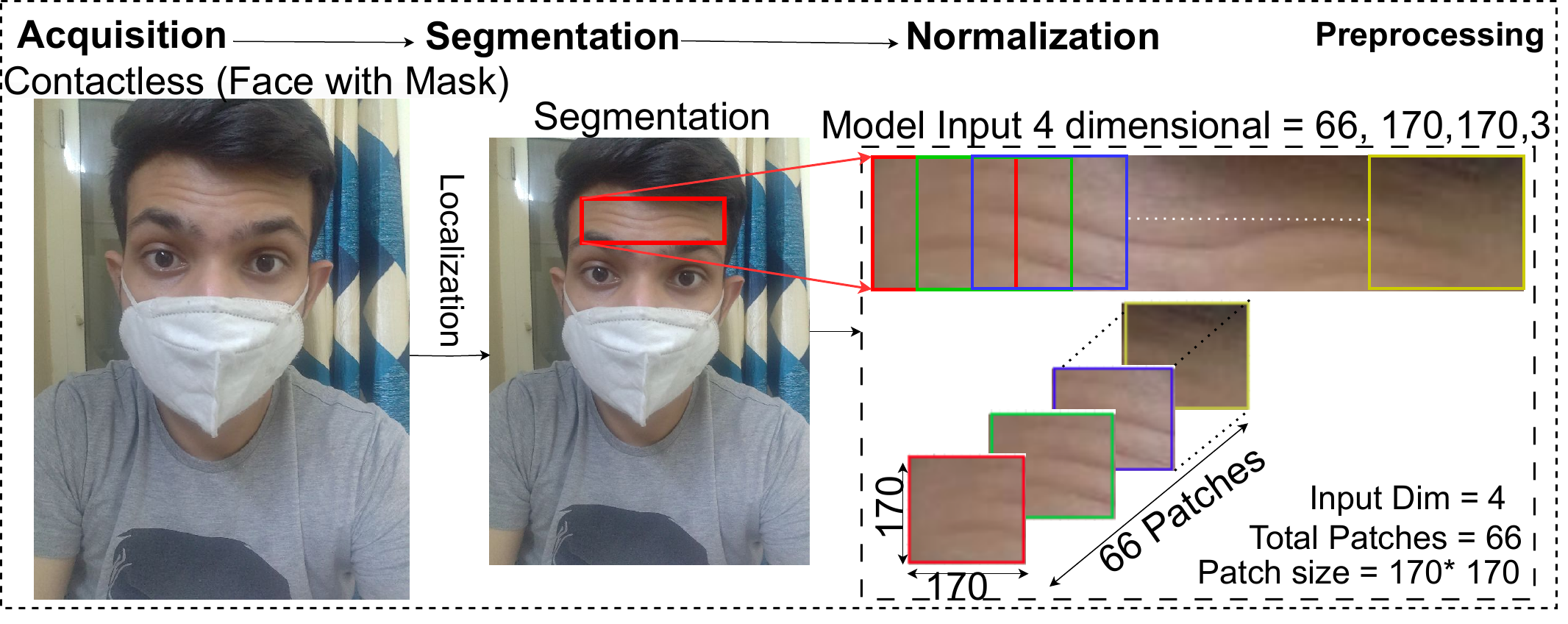}
\caption{A pre-processing procedure to illustrate the utilization of images in the time domain for learning spatio-spatial temporal features. In the first step, we localize and segment the forehead area and then divide it into small-sized overlapped patches with a stride of 5. In the second step, all patches are stacked in the third dimension sequentially to transform them into spatiotemporal format.}
\label{fig:figure2}
\end{figure}

\textcolor{red}{
\IEEEoverridecommandlockouts \IEEEpubid{\makebox[\columnwidth]{TBA \hfill} \hspace{\columnsep}\makebox[\columnwidth]{ }}}
Furthermore, the sequential stacking of these patches introduces a temporal dimension, allowing the model to comprehend the evolution or changes in the features across consecutive frames. Through the systematic processing of sequences comprising interconnected image patches, the model can extract spatio-spatial temporal features. This comprehensive understanding encompasses not only spatio-spatial relationships but also temporal dynamics, providing a holistic perspective on the underlying visual information.
The sequential utilization of small interconnected patches builds on the previously mentioned division of images into patches. This methodology allows the model to concurrently focus on capturing spatial details within each patch while discerning relationships and patterns between neighboring spatial regions through their interconnections. In addition, the sequential stacking of these patches over time introduces a temporal dimension, enabling the model to understand the evolution of features across consecutive frames. This holistic approach facilitates the simultaneous capture of spatial details, spatial relationships, and temporal dynamics, contributing to the model's comprehensive learning of spatio-spatial temporal features. Figure \ref{fig:figure2} illustrates the preprocessing procedure used to extract the forehead image blocks. 

\subsection{FH-SSTNet}
The proposed  FH-SSTNet  is a hierarchical network based on inception-V1 \cite{szegedy2015going} architecture. 
The FH-SSTNet network architecture for forehead recognition consists of two crucial stages: the Backbone and the Head. The Backbone stage serves for initial feature extraction, while the Head stage is dedicated to subsequent analysis and classification of these extracted features.

\subsubsection{Backbone}
The backbone of the FH-SSTNet serves as an important component of forehead recognition. It works as the initial processing unit tasked with extracting essential spatio-spatial temporal features from input forehead creases. The backbone stage efficiently captures both the spatial and temporal characteristics embedded within the image-video-based data.  The FH-SSTNet backbone is composed of five blocks, each of which has sub blocks a,b,c,d,e, and f. The branches of each block are explained as follows:



\begin{figure*}[!ht]
\centering
\includegraphics[width=1\textwidth]{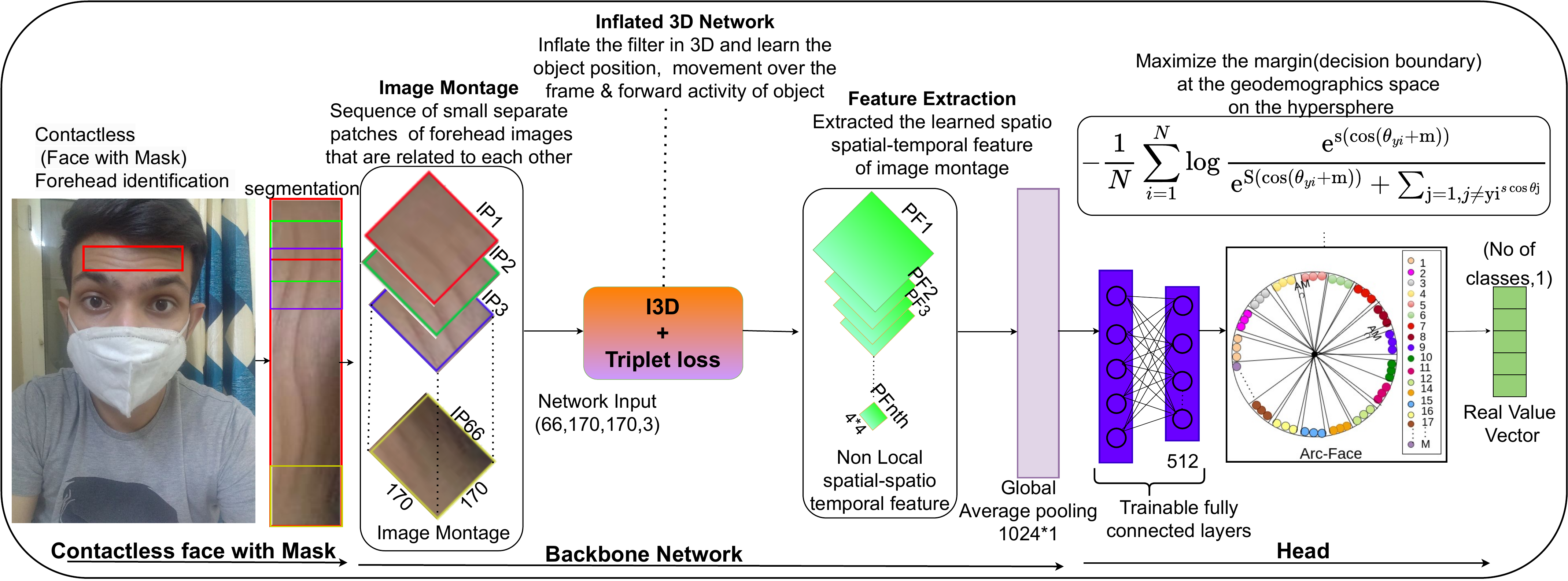}
\caption{Block diagram of the proposed FH-SSTNet for forehead -based person verification. In the first step, a stacked forehead patched into video resulting in 3D representation called a non-local spatio-spatial-temporal relationship is fed into the Backbone network, which generates the non local spatio-spatial temporal features. The following trainable fully connected layer (Head), allows them to process and more discriminate the non-local spatio-spatial temporal feature using Arcloss.}
\label{fig:figure3}
\end{figure*}

\begin{itemize}
    \item \textbf{Block-1} is the foundational stage of the FH-SSTNet. It splits into two paths, branch\_0 and branch\_1, and processes the RGB input data. In branch\_0, a 3D convolutional layer with 32 filters of size (7,3,7) with stride(2,2,2) and the same padding operates on the height, width, and depth dimensions, followed by another convolutional layer with the same filters, reducing the spatial dimensions. Simultaneously, Branch\_1 performs complementary operations. The outputs from both branches are merged, and the resulting feature tensor is down-sampled, yielding a compact 4D output tensor of size $40\times56\times56\times64$.
    \item \textbf{Block-2} comprises branch\_0 and \_1. Branch\_0 features a 3D convolutional layer with 64 filters of size (1,1,1) and a subsequent layer with 81 filters of size 3x3x7. Meanwhile, branch\_1 includes a layer with 81 filters of size $3 \times 7 \times 3$. The outputs from both branches are concatenated and downsampled using a 3D max-pooling layer, resulting in an output of size $40\times28\times28\times162$.
    \item \textbf{Block-3} consists of two sub-blocks: block-3b and block-3c. The outputs of both blocks are concatenated to produce an output feature map of size $40\times14\times14\times480$.
    \item \textbf{Block-4} consists of five sub-blocks: block-4b, block-4c, block-4d, block-4e, and block-4f. Finally, all output features of all sub-blocks are concatenated into one common feature output that contains all features of all blocks of size $20 \times 07 \times 07 \times 832$. 
    \item \textbf{Block-5} comprises two sub-blocks: block-5b and block-5c. Their feature outputs were concatenated to yield features of size $10\times4\times4\times1024$. After average pooling, the output size becomes to 1024. Spatial and temporal features are extracted from block-5 and fed into a global average pooling layer. 

\end{itemize}
The proposed backbone is trained using the triplet loss function, wherein triplets are composed of an anchor, positive sample, and negative sample. The objective is to minimize the distance between the anchor and the positive sample while maximizing the distance between the anchor and the negative sample. This process generates embeddings of size 1024, capturing the distinctive features of the input data. Once the triplet loss is trained, we further refine the network by utilizing these embeddings to train the head using the Arcloss function, as explained below. 

\subsubsection{Head}
The features computed using the backbone of the proposed FH-SSTNet  are fed to the head with two fully connected layers with 1024 and 512 neurons along with Arc loss, to discriminate the 3D spatial and temporal features of the forehead creases as illustrated in Figure \ref{fig:figure3}. 
The Head is trained using ArcLoss function \cite{deng2019arcface} that yields the final features of the dimension $1 \times 512$ that will be used to compute the verification score. 

\subsection{Comparison Score:Cosine similarity}
Given the gallery and probe forehead image, we extract the $1 \times 512$  dimension feature that is further used to compute the verification score using Cosine similarity \cite{nguyen2010cosine}.



\subsection{Implementation Details}
This section presents a comprehensive training methodology comprising of two distinct steps. Initially, we trained our proposed network backbone using the triplet loss function. Subsequently, we refine this training process by focusing on the network head and employing the arc-face loss function to further enhance its performance.  

\textbf{Step 1:} 
We start with the data loader, load each person’s data from the directories, and then sample 100 people per batch and five images per subject to create training batches for the model. 
The training process for a backbone model with triplet loss involves triplet selection. To accomplish this, the process iterates over all images in the same class for each anchor image, identifying the positive and negative images, and computing the cosine distance between the anchor and the other images within the same class (positive sample), as well as images from other classes (negative sample). The construction of triplets involves combining an anchor, a positive image from the same class, and a negative image from a different class. In each iteration, the anchor, positive, and negative images were loaded and their embeddings were computed. These embeddings are then passed through the network to determine the loss and update the network parameters using backpropagation. The triplet loss function was employed to compare the embedding distances between the anchor-positive and anchor-negative pairs. The model is penalized when the distance between the anchor and negative samples is less than the distance between the anchor and positive samples by a margin of 0.5. Consequently, the network learns to embed images such that the embedding of anchor images is closer to positive images and farther from negative ones in the embedding space.

The ADAM optimizer \cite{kingma2014adam} was utilized for gradient descent optimization, with an initial learning rate of 1.0e-5. Each epoch comprised 1000 batches, with each batch containing data from 100 individuals and five images per person. The training process was continued for a maximum of 1000 epochs. Additionally, the margin for triplet selection was initialized to 0.5, with a maximum allowed margin of 1.5. In each training batch, two triplets were processed simultaneously. The hyper-parameters were carefully chosen to ensure effective training and robust performance of our FH-SSTNet in identifying individuals based on their forehead creases despite partial face occlusion due to masks, blurriness, and other challenges present in the dataset.

\textbf{Step 2:} In this stage, we utilize the features learned in Step 1 and further train the head using the Arcloss function. To achieve this, we created a custom layer called Arcloss. The Arcloss layer uses the feature embeddings from the backbone model as input and employs a fully connected layer with weights initialized using the glorot uniform initializer.  The feature embeddings from the backbone model were then passed through a fully connected layer to obtain dense embeddings. These dense embeddings were then passed through the Arcloss layer to compute the final output probability. The model was compiled with 0.5 margin and 30.0 scale hyperparameters and returned along with the base model and complete model (Backbone + Head). With this architecture, we trained the FH-SSTNet with the Arcloss function using a pre-trained backbone model with triplets as feature extractors.

\section{Experiments and Results}
\label{sec:Exp}
In this section, we present the quantitative results of the proposed and existing forehead-verification methods. These results were reported using ISO/IEC 19795 \cite{ISO-19795-1:2021} metrics, specifically, the False Match Rate (FMR) and False Non-Match Rate (FNMR). We also present the results in terms of the True Acceptance Rate (TAR), which is equal to 100 minus the False Non-Match Rate, and the Equal Error Rate (EER), which is computed at an FMR that is equal to the FNMR. Because of the scarcity of existing research on forehead-based user verification, we compared the verification performance of the proposed method with that of the state-of-the-art method reported in \cite{bharadwaj2022mobile} and pre-trained deep CNNs such as ResNet50 \cite{he2016deep}.
\subsection{Dataset}
\begin{figure}[htp]
\centering
\includegraphics[width=0.85\columnwidth]{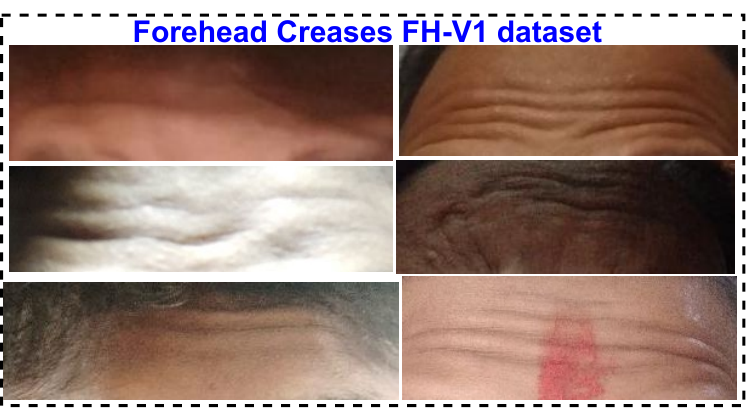}
\caption{Examples  of forehead images corresponding to different subjects from  FH-V1 dataset.}
\label{fig:figure4}
\end{figure}

In this work, we utilized the publicly accessible Forehead Creases version 1 (FH-V1) dataset \cite{bharadwaj2022mobile} to evaluate the performance of the proposed method and state-of-the-art (SOTA) techniques. The FH-V1 dataset comprises 247 subjects, with each individual contributing 20 images captured during two sessions. 
Figure \ref{fig:figure4} shows example images corresponding to different subjects from the FH-V1 dataset.
\subsection{Performance evaluation protocol}
To benchmark the verification performances of the proposed FH-SSTNet and SOTA \cite{bharadwaj2022mobile}, we introduced a common assessment protocol. We concentrated only on closed-set scenarios. Our method processed a stack of image patches as input to learn spatio-spatial-temporal features, with genuine and impostor scores based on these inputs. Each subject contributed approximately 10 to 20 images, which were processed into individual stacks of image patches called 2D temporal sequence, resulting in a total of 247 subjects. Each 2D image is converted into a 2D temporal sequence $(60\times170\times170\times3)$ as shown in Fig \ref{fig:figure2}, which is considered as 1 sample for gallery and probe. Thus, our total genuine samples were calculated as $247\times10\times10 = 24700$ while impostors were determined by $247\times246\times10\times10 = 6076200$.

\begin{table*}

\centering
\caption{}
\label{tab:table1}
\resizebox{1.4\columnwidth}{!}{%
\begin{tabular}{@{}|l|l|ll|@{}}
\toprule
\multirow{2}{*}{Forehead Verification Algorithms} & \multirow{2}{*}{EER(\% )} & \multicolumn{2}{l|}{TMR(\% ) @FMR(\% ) =}            \\ \cmidrule(l){3-4} 
               &       & \multicolumn{1}{l|}{0.1(\% )} & 0.01(\% ) \\ \midrule
ResNet50       & 34.29 & \multicolumn{1}{l|}{5.37}     & 1.58      \\ \midrule
Bharadwaj et al. \cite{bharadwaj2022mobile}           & 5.98  & \multicolumn{1}{l|}{66.48}    & 49.05     \\ \midrule
FH-SSTNet (Our) (Only Backbone)                     & 32.73                     & \multicolumn{1}{l|}{1.15}           & 0.15           \\ \midrule
\textbf{FH-SSTNet (Our) (Backbone and Head)}      & \textbf{1.77}             & \multicolumn{1}{l|}{\textbf{96.58}} & \textbf{91.77} \\ \bottomrule
\end{tabular}%
}
\end{table*}

\subsection{Results and Discussion}
\begin{figure}[htp]
\centering
\includegraphics[width=0.85\columnwidth]{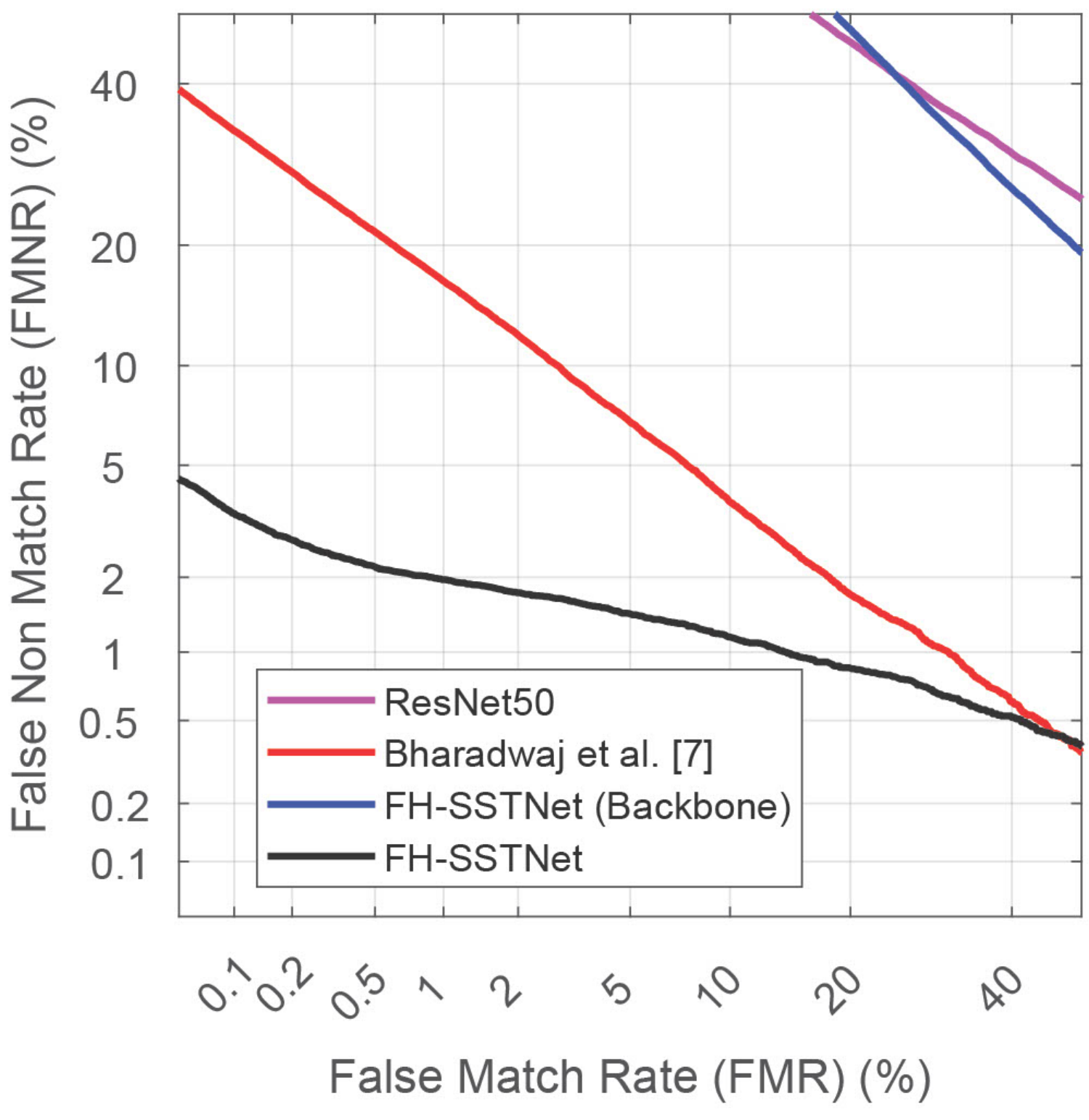}
\caption{DET curve for comparative analysis among four different model. X-axis indicates the false match rate  and  y-axis indicates the false non match rate of forehead creases (FH-V1) dataset.}
\label{Figure:figure5}
\end{figure}

In this section, we perform experiments on the publicly available forehead creases FH-V1 dataset\cite{bharadwaj2022mobile} specifically in a closed-set identification setting. Here, half of the dataset operates as a gallery, while the other half acts as a probe. We perform these experiments in a closed-set scenario without data augmentation. 
Table \ref{tab:table1} presents the performance metrics of various forehead verification algorithms in terms of Equal Error Rate (EER\%), TMR(\%), and FMR(\%) at 0.1\% and 0.01\%. Figure \ref{Figure:figure5} shows the DET curves corresponding to the proposed and existing forehead based person verification methods. ResNet50\cite{he2016deep} achieved an EER of 34.29\% with a TMR and FMR of 5.37\% and 1.58\%, respectively. 
The algorithm proposed by Bharadwaj et al. \cite{bharadwaj2022mobile} exhibited an EER of 5.98\%, with TMR and FMR values of 66.48\% and 49.05\%, respectively. Our proposed algorithm, FH-SSTNet, achieved an EER of 32.73\% with TMR and FMR at 1.15\% and 0.15\%, respectively, when only the backbone was used. When both the backbone and head are used, FH-SSTNet demonstrates a significant improvement, achieving an EER of 1.77\%, with TMR and FMR values of 96.58\% and 91.77\%, respectively. The improved performance of the proposed method can be attributed to the proposed 3D representation in FH-SSTNet, which can enhance the feature representation of forehead biometrics. 
\subsection{Conclusion}
\label{sec:Cocn}
Our proposed FH-SSTNet outperforms other methods, including state-of-the-art approaches. In this work, we introduced an innovative learning model that utilizes 3D spatio-spatial temporal convolution to create detailed representations of forehead images. By incorporating a temporal dimension through sequential patch stacking, we can observe how features evolve across frames. This enabled us to extract spatio-spatial  temporal features, allowing the model to capture both static spatial relationships and their dynamic changes over time. The FH-SST network, which is our novel CNN-based architecture, integrates triplet loss to discern discriminant information. Additionally, the head of the network incorporates Arcloss to further enhance the discriminatory properties of our forehead-crease-based verification method. Extensive experiments on the FH-V1 dataset, which comprised 247 subjects across two sessions, clearly demonstrated the superior performance of our FH-SSTNet. We achieved an Equal Error Rate (EER) of 1.77\% on the Forehead Creases FH-V1 dataset with a False Match Rate (FMR) of 0.1\% and 0.01\% of 96.58 and 91.77 respectively.


{\small
\bibliographystyle{IEEEtran}
\bibliography{ref}
}
\end{document}